\documentclass[runningheads]{llncs}
\usepackage{graphicx}

\usepackage{tikz}
\usepackage{comment}
\usepackage{amssymb}
\usepackage{amsmath} 
\usepackage{color}
\usepackage{booktabs}
\usepackage{multirow}
\usepackage{array}

\usepackage[accsupp]{axessibility}  

\newcolumntype{x}[1]{>{\centering\arraybackslash\hspace{0pt}}p{#1}}

\begin{document}
\pagestyle{headings}
\mainmatter
\def\ECCVSubNumber{}  

\title{Action-based Contrastive Learning for Trajectory Prediction} 



%
\author{Marah Halawa \and
Olaf Hellwich \and
Pia Bideau}
\authorrunning{M. Halawa et al.}
%
\institute{Excellence Cluster Science of Intelligence, Technische Universitat Berlin 
\email{\{marah.halawa,olaf.hellwich,p.bideau\}@tu-berlin.de}}

\maketitle

\begin{abstract}
Trajectory prediction is an essential task for successful human-robot interaction, such as in autonomous driving. 
In this work, we address the problem of predicting future pedestrian trajectories in a first-person view setting with a moving camera.
To that end, we propose a novel action-based contrastive learning loss, that utilizes pedestrian action information to improve the learned trajectory embeddings. 
The fundamental idea behind this new loss is that trajectories of pedestrians performing the same action should be closer to each other in the feature space than the trajectories of pedestrians with significantly different actions. 
In other words, we argue that behavioral information about pedestrian action influences their future trajectory.
Furthermore, we introduce a novel sampling strategy for trajectories that is able to effectively increase negative and positive contrastive samples. Additional synthetic trajectory samples are generated using a trained Conditional Variational Autoencoder (CVAE), which is at the core of several models developed for trajectory prediction.
Results show that our proposed contrastive framework employs contextual information about pedestrian behavior, \textit{i.e.} action, effectively, and it learns a better trajectory representation. 
Thus, integrating the proposed contrastive framework within a trajectory prediction model improves its results and outperforms state-of-the-art methods on three trajectory prediction benchmarks \cite{Rasouli2019PIE,rasouli2017ICCVW,malla2020titan}.
\end{abstract}
%
%
\section{Introduction}
Predicting the future trajectories of pedestrians is an important task in many applications, such as in social robot interaction and autonomous driving. Typically, the future trajectory of an agent/pedestrian is predicted based on its own past movement history \cite{doi:10.1177/0278364920917446}. Nonetheless, integrating additional information is possible, such as the trajectories of surrounding agents \cite{slstm,sconv}, or visual scene data \cite{Sadeghian_2019_CVPR}. 
When the surrounding agents in the scene are cars or robots, modeling the motion information based on past trajectories only is a reasonable way to solve the task. However, in this work, we argue that when other agents in the scene are pedestrians, then limiting the information used for prediction to past trajectories is not sufficient. In those cases additional information about pedestrian behavior (\textit{e.g.} action) plays an important role for predicting their future trajectory.
For example, the future trajectories of a pedestrian who is walking while texting on a phone could be different from a pedestrian carrying an object or pushing a baby stroller even if they have the same previous observed trajectories, and the same end goal.

\begin{figure}
\centering
\includegraphics[width=\linewidth]{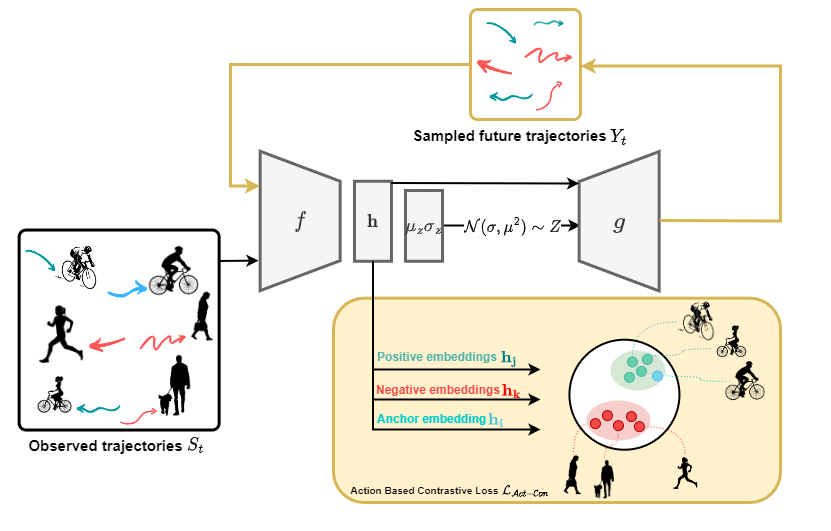}
\caption{Overview of our action-based contrastive learning framework during training phase. The contrastive loss $\mathcal{L}_\text{Act-Con}$ gets as input both positive (green) and negative (red) embeddings $h$ for an anchor (blue). The positive and negative samples are the samples other than the anchor in the batch, as well as the synthetic samples from the CVAE. The parts shown in yellow refer to our novel action-based contrastive learning framework. It is worth mentioning that the action-based contrastive loss illustrated in this figure updates only the weights of encoder $f$, and it is jointly optimized with $\mathcal{L}_{traj}$ that updates both encoder $f$ and decoder $g$. $\mathcal{L}_{traj}$ is not shown in the figure.}
\label{fig1}
\end{figure}

In this work, we study the influence of observed pedestrians’ actions on their predicted trajectories. We propose a novel contrastive learning loss called \textit{Action-based Contrastive Loss}. This novel loss is employed as a regularizer to the main trajectory prediction loss. The action-based contrastive loss encourages the trajectory embeddings of agents performing the same action (called positive samples) to come closer to each other in the feature space, and the embeddings of trajectories observed while performing different actions (called negative samples) far away from each other. For instance, the representations of trajectories of walking pedestrians are encouraged to become closer in the feature space, but farther from the representations of trajectories of pedestrians riding bikes or standing, as illustrated in Fig.~\ref{fig1}.

Contrastive learning losses, including ours (action-based contrastive loss), utilize a mechanism called negative sampling/mining, which aims to choose the samples that are deemed different and therefore their corresponding features are driven farther in the embedding space. In our case, the negatives are trajectories of pedestrians that have different actions. Commonly used negative sampling techniques include choosing all other samples from the same mini-batch~\cite{simclr} or from a fixed-size memory bank~\cite{moco}.
Nevertheless, while these mechanisms prove effective on natural imaging datasets, we find they do not provide similarly high gains on trajectory datasets. We conjecture that this is due to the higher variation in visual data compared to trajectory data, and most importantly, to the larger sizes of imaging datasets, \textit{e.g.} Imagenet \cite{deng2009imagenet} contains 1.6M images compared to PIE \cite{Rasouli2019PIE} that contains 738,970 trajectory samples.
This results in limited numbers of negative samples, an issue that becomes more evident when conditioning samples by class information, \textit{e.g.} action or behavior. Few works attempt to address this issue via designing special heuristics for negative mining \cite{liu2021social,Wu_2017_ICCV,Harwood_2017_ICCV}. Alternatively, in this work, we propose to utilize the data distribution learned by a Conditional Variational Auto-Encoder (CVAE) \cite{NIPS2015_8d55a249}. This avoids designing special heuristics for negative mining. While this form of sampling may be utilized to create negative samples only, we employ it to create both positive and negative samples. This is possible due to the different definition of our contrastive loss compared to the traditional Noise Contrastive Estimation loss (NCE loss); the notion of positive/negative in our case is tied to the different classes of action in the data. As explained above, the samples that belong to the same action class are positives from the point of view of this class, and other samples are negative.

\noindent \textbf{Contributions.} Our main contributions in this paper are as follows:
\begin{itemize}
\item A novel contrastive loss, called action-based contrastive loss, which provides the model with additional information about the action of an agent by guiding the development of the embedding space for trajectories during learning. 
\item A novel sampling/mining technique that utilizes the latent trajectory distributions learned by CVAEs, circumventing the need to design special mechanisms based on heuristics.
\end{itemize}
Our proposed contrastive learning framework improves the performance results on three first-person view trajectory prediction benchmarks. It also provides evidence that utilizing agent behavior information, in the form of action type in this case, is beneficial for trajectory prediction, aligning with ~\cite{malla2020titan}. However, our proposed learning framework requires action information only during training.

\section{Related Work}
\textbf{Multi-modal trajectory prediction:}
A human can reach a desired location following many possible trajectories. Therefore, multiple works utilize multi-modal trajectory models, instead of predicting a single-path solution.
Lee \textit{et al.}~\cite{lee2017desire} proposed multi-modal trajectory model by incorporating samples from the Gaussian distribution of a trained conditional variational autoencoder (CVAE) into a long short-term memory encoder-decoder (LSTMED) model.
Mangalam \textit{et al.} \cite{DBLP:conf/eccv/MangalamGALAMG20} predict the multi-modal trajectory of an agent by modeling three factors: the desired endpoint goal, the social interaction with other agents in the scene, and the planned trajectories with respect to the environmental constraints in the scene. Similarly, their model is based on CVAE, which takes as input both the encodings of the past trajectory and of the endpoint goal. 
Sadeghian \textit{et al.} \cite{Sadeghian_2019_CVPR} additionally include the past/observed trajectories of all agents for future trajectory prediction. To provide additional context information, top view images are incorporated. The distribution over feasible future paths is modeled for each agent using LSTM-based GAN module.
Similarly, Yao \textit{et al.} \cite{Yao:2021aa} predict trajectories conditioned on an estimated goal using a bi-directional RNN decoder.
While our method has the potential of being added to any trajectory prediction method, we base our contrastive framework on BiTraP \cite{Yao:2021aa}, in the first-person view setting.

\noindent\textbf{Using human actions to improve trajectory prediction:}
In literature, many works employed video data to predict human activities \cite{pareek2021survey}. Montes \textit{et al.}~\cite{Montes_2016_NIPSWS} used a 3D-CNN as a feature extraction network then pass the learned representation to an RNN to exploit the time component in video data effectively. Ma \textit{et al.}~\cite{Ma2016LearningAP} improved the performance of LSTMs in human activity prediction by implementing ranking losses that penalize the prediction model on inconsistency in prediction scores from the sequence frames. Liang \textit{et al.}~\cite{liang2019peeking} predicted a pedestrian's future trajectory simultaneously with future activities in a multi-task learning scheme. Rasouli \textit{et al.}~\cite{Rasouli2019PIE}, studied the influence of an estimated pedestrian intention on the predicted trajectory by combining the intention representation with the observed trajectory coordinates, then used this representation as input to the decoder. 
Malla \textit{et al.}~\cite{malla2020titan} incorporates pedestrian action information with a trajectory prediction model. They require this information as prior information and learn a joint representation for both observed trajectory and pedestrian action. 
In this work, we also highlight the importance of analyzing the pedestrian's behavior and action in the prediction of their future trajectory. However, we propose to incorporate action information only during training using a novel action-based contrastive loss. 

\noindent\textbf{Contrastive trajectory prediction:}
Contrastive learning is a representation learning approach, first proposed by \cite{van2018representation}. This approach encourages similar high-dimensional input vectors to be mapped closely to each other in a lower-dimensional embedding manifold, and the dissimilar ones are mapped far away from each other. 
Contrastive learning has been applied in several unsupervised \cite{van2018representation,CPC2,simclr,byol,barlow,swav,moco,simsiam,nnclr,jahanian2022generative} and supervised \cite{NEURIPS2020_d89a66c7} representation learning methods.
Recently, only few works applied contrastive learning to trajectory prediction in a multi-agent setting. The flexibility of defining a contrastive loss by using positive and negative samples addresses the shortage problem in critical and challenging scenarios in training datasets. Such rare scenarios are necessary for the model, as the agent could face these in the real-world. Makansi \textit{et al.}~\cite{Makansi_2021_ICCV} utilize this idea by separating the hard and critical samples in the feature space that do not satisfy some certain favorable criterion far away from the positive easy samples. Liu \textit{et al.} \cite{liu2021social} proposed a social sampling strategy that relies on augmenting negative samples with prior knowledge about undesired scenarios in the multi-agent setting. Both methods use the contrastive loss as a weighted combination to the future trajectory forecasting loss, which may be the mean squared error (MSE) or negative log-likelihood (NLL).
Our method follows this family of algorithms, and uses a novel action-based contrastive loss to add context information about pedestrian actions to the trajectory prediction model. 

\noindent\textbf{Supervised contrastive loss:} 
Khosla \textit{et al.}~\cite{NEURIPS2020_d89a66c7} proposed a supervised contrastive loss that is a generalization of the Triplet loss~\cite{hoffer2015deep}. In this supervised contrastive loss for each anchor there are more than one positive sample, in addition to many negative samples. There are two major differences between our proposed action-based contrastive loss and the supervised contrastive loss used in~\cite{NEURIPS2020_d89a66c7}. First, they employ the supervised contrastive loss to \textit{replace} the cross-entropy loss for training the image classifier using image labels. However, we utilize the contrastive loss to \textit{regularize} the trajectory prediction loss, which may be MSE or NLL. 
Second, due to the differences between the nature of datasets we use in this paper and the image data used in \cite{NEURIPS2020_d89a66c7}, it is simpler to extract many positive and negative samples from a large dataset, such as ImageNet \cite{deng2009imagenet}. However, in first-person view trajectory prediction datasets, the number of pedestrians with same actions is limited, therefore we address this with a novel sampling process from a CVAE trained to predict trajectories based on observing a short past trajectory. This CVAE predictive model ensures consistency between observed and predicted trajectories. Thus, it allows sampling additional positive and negative samples that belong to specific actions. Using this novel sampling technique avoids designing hard negative mining techniques, which use heuristics, as in ~\cite{Domain_ad,kim2021selfreg} for domain adaptation.

\section{Methodology}
In this section, we present our method for the task of pedestrian trajectory prediction, 
that focuses on integrating contextual information such as actions for more reliable future predictions.
We address this by employing an action-based contrastive loss that enhances the trajectory prediction model with action information.
\subsection{Problem Formulation}
For each pedestrian we have an observed past trajectory $S_t = [s_1,..., s_{t-1}, s_t]$ at time $t$, and we predict a future trajectory $Y_t = [y_{t+1}, y_{t+2}, ..., y_{T}]$, where $s$ and $y$ are bounding box coordinates
for the observed and predicted trajectories, respectively. $T$ is the maximum predicted trajectory time length in the future. 
In addition, we also have for each trajectory the action class information $a$, where the set of available actions $a \in \{a_1, a_2, ..\}$ may vary across different datasets. 
Then in the training data, we assume there are $N$ different training samples, where for each sample $i \in [1,..,N]$, we know $S^i$, $Y^i$, and $a^i$.
Finally, we process the dataset samples in mini-batches, where each batch contains $B$ samples. 

\subsection{Multi-modal Trajectory Prediction} \label{sect:traj_pred}
We follow the commonly used approach of an encoder-decoder prediction model, where an encoder $f$ learns the representation $h$ given an observed trajectory $S_t$ as an input, then a decoder $g$ uses the representation $h$ together with a sampled latent variable $z$ to predict the future trajectory $Y_{t}$. 
We employ a standard long-short term encoder-decoder model (LSTMED)~\cite{lee2017desire}. 
In fact, we extend on the bi-directional version of LSTMED, proposed in Yao \textit{et al.}'s BiTraP~\cite{Yao:2021aa}.
The possibility to draw multiple future trajectories for each observed trajectory is achieved with a CVAE, which is a non-parametric model, that learns the distribution of target trajectory through a stochastic latent
variable. The distribution learned by the CVAE is essential for our proposed contrastive framework, which we explain below. As a trajectory prediction loss function $\mathcal{L}_{traj}$, the Best-of-Many (BoM) L2-loss~\cite{bhattacharyya2018accurate} between predicted and target trajectory is used. 
It is noteworthy that we do not restrict our proposed framework, explained below, to these choices of model architectures or loss functions; we adopt standard and effective techniques to study its influence on predicted trajectories. The essential factor for our learning framework is that the predicting future trajectory model is based on CVAE, similar to trajectory prediction models in~\cite{Yao:2021aa,DBLP:conf/eccv/MangalamGALAMG20}.
\subsection{Action-based Contrastive Learning Framework} \label{sect:ance}
In order to enhance the model with contextual information about the pedestrian actions, we propose a novel loss that is called action-based contrastive loss, which acts as a regularizer for the trajectory prediction loss, and they jointly train the trajectory prediction model. The proposed action-based contrastive loss is based on a novel action-based sampling strategy shown in Fig.~\ref{fig1}. We first describe the proposed contrastive loss in the simple case, without including additional samples from the CVAE distribution, and we generalize it later. 

\subsubsection{Action-based Contrastive Loss}
Let $B$ be the number of samples within a batch. For each observed past trajectory $S^i$ where $i \in \{1,..,B\}$, called the anchor, there exists multiple positive and negative samples. 
The positive samples $S^{i+}$ are the trajectories that have the same action class as the anchor, which are denoted by ${S^i}''$. Moreover, we also add an augmented version ${S^i}'$ of the anchor trajectory as a positive sample, following~\cite{liu2021social}, which is created by adding small white noise $\epsilon$ to the bounding box coordinates of the anchor trajectory.

Formally: 
\begin{center}
${S^i}' = \left \{ S^i + \epsilon \right \}$ \\
${S^i}'' = \left \{ S^{j}  \right \}; \text{where} \; 0<j<B, a^j = a^i, i \neq j$ \\
$S^{i+} = {S^i}' \cup  {S^i}''$ 
\end{center} 
Negative samples $S^{i-}$ are trajectories belonging to a different action class than the anchor.
\begin{center}
$S^{i-} = \left \{ S^{k}  \right \}; \text{where} \;  0<k<B, a^k \neq a^i, i \neq k$
\end{center} 
Afterwards, all batch samples $\left \{ S^i \right \}_{i=1}^{B}$ are processed by the model encoder $f$ to produce their hidden representations $\left \{ \mathbf{h}^i \right \}_{i=1}^{B}$. Assuming $M$ positive samples and $K$ negative samples in the batch, with $B=M+K$. The proposed loss is calculated as follows:
\begin{equation} \label{eq_actcon}
\begin{split}
\ell_\text{Act-Con} &= -\frac{1}{B}\sum_{i=1}^B \log \frac{\sum^M_{j=1, j\neq i, a^j = a^i} \exp(sim(\mathbf{h}^i, \mathbf{h}^j) / \tau)}{\sum^K_{k=1,k\neq i} \exp(sim(\mathbf{h}^i, \mathbf{h}^k) /\tau)} \\
\mathcal{L}_\text{Act-Con} &= \frac{1}{N/B}\sum^{N/B}\ell_\text{Act-Con}
\end{split}
\end{equation}
where $sim$ is the similarity between the vector representations of the samples, for which we use the dot-product. $\tau$ is the temperature hyperparameter.
The above loss function encourages the embeddings $\mathbf{h}^i$ of positive sample trajectories to be closer to each other in the embedding space, and far away from the embeddings of negative samples. 
The complete loss function sums both the trajectory prediction loss $\mathcal{L}_{traj}$ and the action-based contrastive loss $\mathcal{L}_\text{Act-Con}$: 
\begin{equation} \label{eq_final}
\mathcal{L}_{final} = \mathcal{L}_{traj} + \beta \mathcal{L}_\text{Act-Con}
\end{equation}
where $\beta$ is a hyper-parameter that controls the contribution of action-based contrastive loss.
It is worth mentioning that additional behavioral information such as pedestrian's action class is only needed during training. However, during inference, the model only takes the observed trajectory as input to predict the future trajectory.

\subsubsection{Action-based Synthetic Trajectory Sampling}
\begin{figure}
\centering
{\includegraphics[width=.3\linewidth]{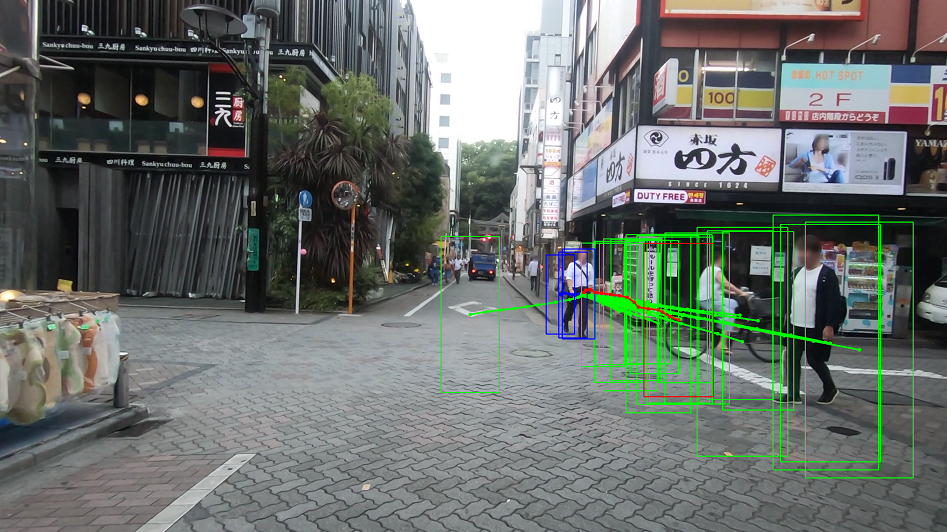}} 
{\includegraphics[width=.3\linewidth]{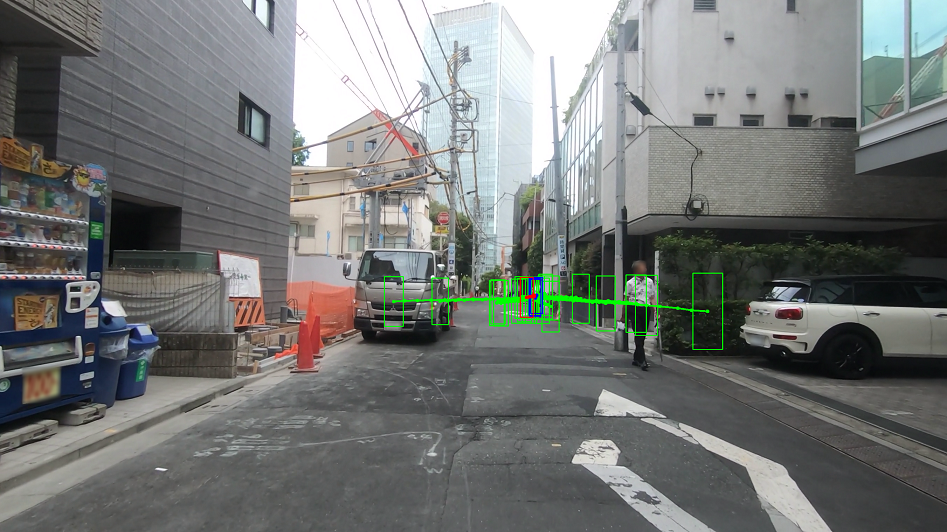}}
{\includegraphics[width=.3\linewidth]{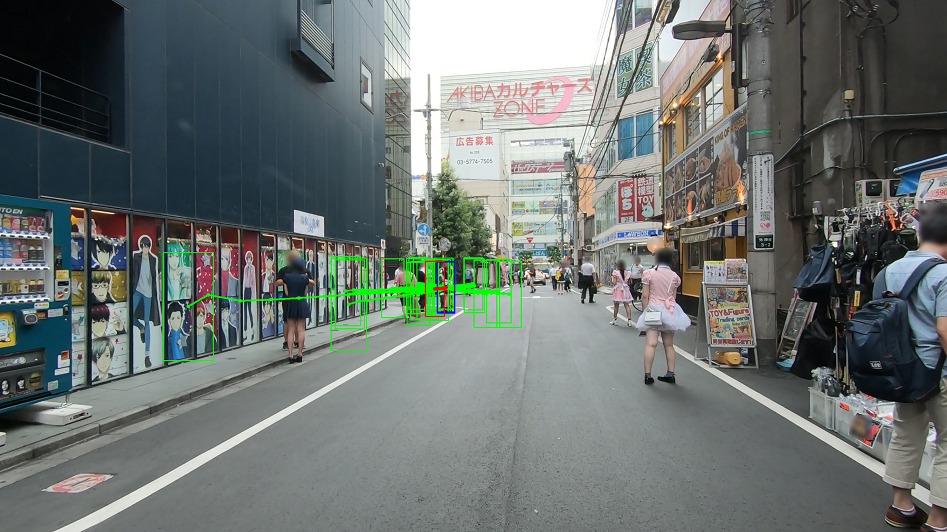}}\\
{\includegraphics[width=.3\linewidth]{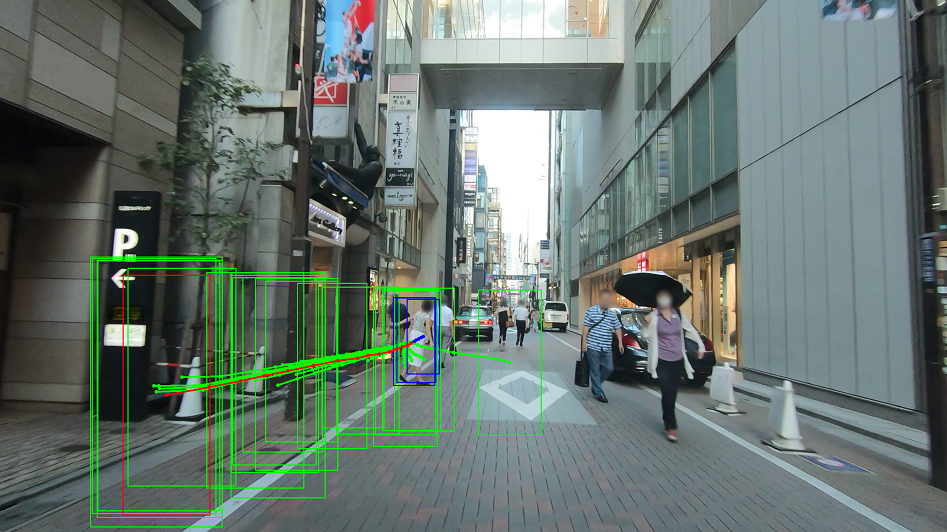}} 
{\includegraphics[width=.3\linewidth]{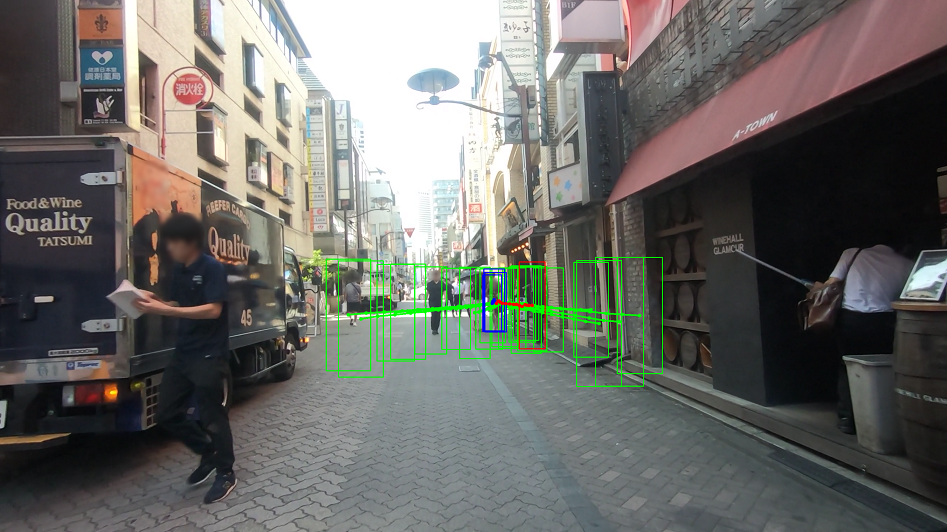}} 
{\includegraphics[width=.3\linewidth]{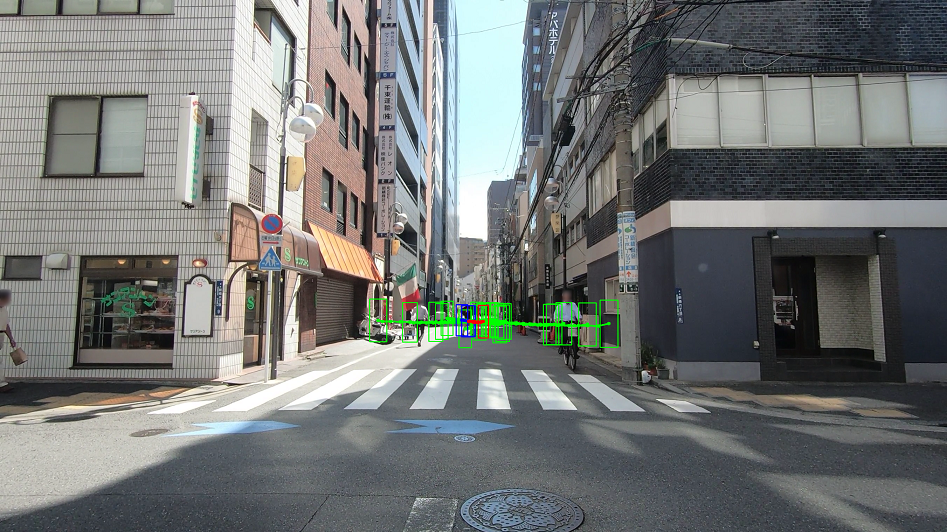}} 
\caption{Examples from TITAN dataset showing the multi-modality in the trajectory prediction space. CVAE is able to predict multiple feasible future trajectories (Green bounding boxes), conditioned on previously observed trajectories (Blue bounding boxes). The red bounding boxes refer to the ground truth future trajectories.}
\label{fig2}
\end{figure}

The above loss formulation assumes no additional synthetic samples, \textit{i.e.} it considers observed trajectories in the batch only.
However, due to the relatively limited sizes of trajectory datasets, and the shortage of diversity in action classes in captured scenes, commonly used negative sampling techniques may not be sufficient.
Those include sampling from the same mini-batch~\cite{simclr} or from a fixed-size memory bank~\cite{moco}.
More comprehensive negative and positive samples, from various behavior scenarios are rather needed. Therefore, we extend training samples by drawing trajectories from the distribution learned by the generative Conditional Variational Autoencoder (CVAE) model. 
CVAE is a generative model that introduces a stochastic latent variable $Z$ in order to learn the distribution of target future trajectory $P\left ( Y^i|S^i,Z \right )$. This distribution is conditioned on the input observed trajectories $S^i$, and the stochastic latent variable $Z$. Thus, the model is able to predict \textit{multiple} feasible trajectories $Y^i$ given the input $S^i$. 
We assume the latent variable following a Gaussian distribution $Z \sim N \left ( \mu_Z ,\sigma_Z^{2} \right )$, and we train the CVAE to capture this distribution. 
Afterwards, the training dataset is extended by sampling from the Gaussian latent space multiple times, and passing samples through the decoder $g$ to effectively predict different feasible future trajectories conditioned on an observed trajectory. The conditioning on the observed trajectory ensures a consistent behavior in the predicted future trajectory. This behavior is captured in both the continuity of the trajectory as well as the identical action class in both observed and future trajectories.
We employ the same encoder-decoder trajectory prediction model explained above in Sec.~\ref{sect:traj_pred}, which is a CVAE that predicts multiple feasible future trajectories, as the example in Fig.~\ref{fig2} shows. 
This sampling strategy is illustrated in Fig.~\ref{fig1}, and it has the advantage that it avoids designing heuristics for negative sample mining techniques, as mentioned before. 
The intuition behind this sampling strategy is that the encoder-decoder CVAE model is capable of generating future trajectories with the same behavior of the observed trajectory. Since it is trained to predict the future trajectory of an observed trajectory, then it captures the characteristics of the observed trajectory.  

Let $\left \{ Y^{i,l}\right \}_{l=1}^{L}$ be the multiple predicted trajectories for an observed trajectory $S^i$. Here, $Y^{i,l}$ is sampled from $P\left ( Y^{i,l}|S^i,Z \right )$, and $L$ is the number of times we sample a different $Z$ from the normal distribution $N\left( \mu_Z ,\sigma_Z^{2} \right)$. Given these synthetic trajectory samples, the set of positive samples for trajectory $S^i$ are then reformulated as follows:
\begin{center}
$S^{i+}_{1:t} = \left \{{S^i}'_{1:t} \right \} \cup \left \{{S^i}''_{1:t} \right \} \cup \left \{Y^{j,l}_{t+1:T}\right \}_{l=0}^L$ \\
\end{center}
where $i,j \in {1,..,B}$ and $a^l = a^i$ and $a^j = a^i$. And the negative samples are reformulated as follows:
\begin{center}
$S^{i-}_{1:t} = \left \{ S^{k}_{1:t}  \right \} \cup \left \{ Y^{k,l}_{t+1:T}  \right \}_{l=0}^L$ 
\end{center}
where $i, k \in {1,..,B}$ and $i \neq k$ and $a^i \neq  a^k$ and $a^i \neq  a^l$ and $a^k = a^l$. In words, the synthetic samples created for sample $S^k$, which we denote $Y^{k,l}$, are considered negative from the point of view of sample $S^i$. These synthetic samples have the same action class of sample $S^k$, hence denoted $a^k = a^l$.

The described action-based synthetic trajectory sampling strategy changes the sets of positive and negative samples used in creating training batches. However, the proposed contrastive loss equation Eq.~\ref{eq_actcon} remains the same, only $M$ and $K$ are affected.

\section{Experiments}
In this section, we present the evaluation results of our method on three first-person-view trajectory prediction datasets \cite{Rasouli2019PIE,rasouli2017ICCVW,malla2020titan}. First, we describe the used datasets. Then, we provide an overview of the experimental setup and used evaluation metrics. Finally, we discuss our results and findings. 

\subsection{Datasets}
We evaluate our method on first-person view datasets. In this domain, the Pedestrian Intention Estimation (PIE) \cite{Rasouli2019PIE} and the Joint Attention for Autonomous Driving (JAAD) \cite{rasouli2017ICCVW} datasets are the most commonly used benchmarks in literature.
The PIE dataset provides 293,437 annotated frames, containing 1,842 pedestrians with behavior annotations such as walking, standing, crossing, looking, etc. Since a pedestrian could be “walking” and “looking” at the same time, for example, then a pedestrian could have multiple behavior labels in a single frame. Therefore, we only use two classes “walking” and “standing”, which are exclusive. We use the same train and test splits in \cite{Rasouli2019PIE}. On the other hand, the JAAD dataset provides 82,032 annotated frames, containing 2,786 pedestrians, 686 of them have behavior annotations. Similar to the PIE dataset, we only use for JAAD dataset two classes “walking” and “standing”, which are exclusive. We use the same train and test splits in \cite{rasouli2017ICCVW}. 

We also use a third dataset named TITAN~\cite{malla2020titan}, which contains more action classes compared to PIE and JAAD. 
TITAN provides 75,262 frames with 395,770 pedestrians with multiple action labels organized in five hierarchical contextual activities, such as individual atomic actions, simple scene contextual actions, complex contextual actions, transportive actions, and communicative actions.
For the same reason of not having multiple labels for each pedestrian, we use individual atomic actions labels for the TITAN dataset. The atomic action labels describe the primitive action, and are categorized into 9 labels (sitting, standing, walking, running, bending, kneeling, squatting, jumping, laying down).

\begin{table}
\begin{center}
\caption{The quantitative results on \textbf{PIE} and \textbf{JAAD} datasets. The evaluation metrics are reported for different prediction lengths in \textit{squared pixels}. ABC+ is our proposed action-based contrastive framework with sampling from a learned CVAE. BiTraP is a baseline trajectory prediction model without adding any contrastive loss. The other baseline results are obtained from \cite{Yao:2021aa}. Lower is better.}
\label{table:pie_jaad}
\begin {tabular}{ lx{0.7cm}x{0.7cm}x{0.7cm}  x{1cm}x{1cm} | x{0.7cm}x{0.7cm}x{0.7cm}  x{1cm}x{1cm}}
\toprule
\multirow{3}{*}{Method} &  \multicolumn{5}{c|}{\textbf{PIE}} & \multicolumn{5}{c}{\textbf{JAAD}} \\
\cline{2-11}
& \multicolumn{3}{c}{ADE} & \multicolumn{1}{c}{C-ADE}& \multicolumn{1}{c|}{C-FDE} & \multicolumn{3}{c}{ADE} & \multicolumn{1}{c}{C-ADE}& \multicolumn{1}{c}{C-FDE}\\
 & 0.5&1.0&1.5&  1.5&  & 0.5&1.0&1.5&  1.5&\\
\hline
Linear \cite{Rasouli2019PIE}               & 123 & 477 & 1365 & 950 & 3983 & 233 & 857 & 2303 & 1565 & 6111 \\
LSTM \cite{Rasouli2019PIE}                 & 172 & 330 & 911 & 837  & 3352 & 289 & 569 & 1558 & 1473 & 5766 \\
B-LSTM \cite{Bhattacharyya2018LongTermOP}  & 101 & 296 & 855 & 811 & 3259 & 159 & 539 & 1535 & 1447 & 5615 \\
FOL-X \cite{Yao2019EgocentricVF}           & 47 & 183 & 584 & 546  & 2303 & 147 & 484 & 1374 & 1290 & 4924 \\
$PIE_{traj}$ \cite{Rasouli2019PIE}         & 58 & 200 & 636 & 596 & 2477 & 110 & 399 & 1280 & 1183 & 4780 \\
\hline
BiTraP \cite{Yao:2021aa}                   & 23 & 48 & 102 & 81 & 261 & \textbf{38} & 94 & 222 & 177 & 565 \\
\hline
ABC+                                       & \textbf{16} & \textbf{38} & \textbf{87} &  \textbf{65} & \textbf{191} & 40 & \textbf{89} & \textbf{189} & \textbf{145} & \textbf{409} \\
\bottomrule
\end{tabular}
\end{center}
\end{table}

\subsection{Experimental Setup}
We use the same setup for all datasets, where we observe 0.5 seconds and predict 0.5, 1.0, and 1.5 seconds, following~\cite{Rasouli2019PIE,Yao:2021aa}. The predicted trajectories have two forms: bounding boxes coordinates and centers, that are evaluated separately. PIE and JAAD datasets are both annotated at 30Hz frequency, therefore we observe 15 frames and predict 45 frames. However, TITAN dataset is annotated at 10HZ sampling frequency. Thus, we observe 5 frames and predict 15 frames.

\subsubsection{Implementation details.} We use 256 as the size for all hidden layers in the encoder-decoder model that is detailed in Sec.~\ref{sect:traj_pred}. It is noteworthy that we implement the loss in Eq.~\ref{eq_actcon} using the efficient matrix-form (especially on GPU machines), instead of performing expensive pairwise computations. We train the model on all datasets with Adam optimizer~\cite{kingma2014adam} using a batch size of 128 and a learning rate of 0.001. 
On training datasets, we perform hyper-parameter tuning for $\beta$ (Eq.~\ref{eq_final}). We achieve our best results using $ \beta = 0.75$ for all datasets. 
\subsubsection{Evaluation metrics.}
Following the commonly used evaluation protocols in literature~\cite{Rasouli2019PIE,Yao:2021aa,Stepwise}, we use the following evaluation metrics: i) Bounding box Average Displacement Error (ADE), ii) Bounding box Center ADE (C-ADE), iii) Bounding box Final Displacement Error (FDE), and iv) Bounding box Center FDE (C-FDE). All are computed in squared pixels.
The bounding box ADE is the mean square error (MSE) for all predicted trajectories and ground-truth future trajectories. This error is calculated using the bounding box upper-left and lower-right coordinates. However, in C-ADE, otherwise called C-MSE, the error is calculated using the centers of the bounding boxes.
Bounding box FDE, otherwise called FMSE, is the distance between the destination point of the predicted trajectory and of the ground truth at the last time step. FDE is also calculated using the bounding boxes coordinates. Finally, C-FDE or C-FMSE is the mean squared error between the centers of final destination bounding boxes.
\begin{table}
\begin{center}
\centering
\caption{The quantitative results on \textbf{TITAN} dataset. The evaluation metrics are reported for observing 15 time steps and predicting 45 time steps of trajectories in \textit{squared pixels}. ABC+ is our proposed action-based contrastive framework with sampling from a learned CVAE. BiTraP is a baseline trajectory prediction model without adding any contrastive loss. Lower is better.}
\label{table:titan1}
\begin{tabular}{ l x{0.7cm}x{0.7cm}x{0.7cm} x{1cm} x{1cm} } 
\toprule
\multirow{2}{*}{Method} &  \multicolumn{3}{c}{ADE} & \multicolumn{1}{c}{C-ADE}& \multicolumn{1}{c}{C-FDE} \\
 & 0.5&1.0&1.5&  1.5&  \\
\hline
BiTraP \cite{Yao:2021aa} & 194 & 352 & 658 & 498 & 989 \\
\hline
ABC+                     & \textbf{165} & \textbf{302} & \textbf{575} & \textbf{434} & \textbf{843} \\
\bottomrule
\end{tabular}
\end{center}
\end{table}
\begin{table}
\begin{center}
\caption{The quantitative results on \textbf{TITAN} dataset. The evaluation metrics are reported for observing 10 time steps and predicting 20 time steps of trajectories in \textit{pixels}. ABC+ is our proposed action-based contrastive framework with sampling from a learned CVAE. The other baseline results are obtained from \cite{malla2020titan}. Lower is better.}
\label{table:titan2}
\small \begin {tabular}{lx{1cm}x{1cm}}
\toprule
Method & ADE & FDE \\
\hline
Social-LSTM \cite{slstm}            & 37.01 & 66.78 \\
Social-GAN \cite{Gupta2018SocialGS} & 35.41 & 69.41 \\
Titan-vanilla \cite{malla2020titan} & 38.56 & 72.42 \\
Titan-AP \cite{malla2020titan}      & 33.54 & 55.80 \\
\hline
ABC+                                & \textbf{30.52} & \textbf{46.84} \\
\bottomrule
\end{tabular}
\end{center}
\end{table}
\subsubsection{Baselines.} The trajectory prediction model trained with our proposed \underline{A}ction-\underline{B}ased \underline{C}ontrastive framework (loss and sampling strategy) is indicated by (ABC+). First, we evaluate the performance of our action-based contrastive framework by comparing its results to the original BiTraP trajectory prediction model~\cite{Yao:2021aa} on all datasets, \textit{i.e.} without adding our contrastive loss. This baseline aims to highlight the gains obtained by our proposed contrastive framework. BiTraP had previously achieved state-of-the-art on PIE and JAAD datasets.
Additionally, for PIE and JAAD datasets, we compare our results with PIEtraj \cite{Rasouli2019PIE}, FOL-X \cite{Yao2019EgocentricVF}, B-LSTM \cite{Bhattacharyya2018LongTermOP}, LSTM \cite{Rasouli2019PIE}, and Linear \cite{Rasouli2019PIE} trajectory prediction models.
On the TITAN dataset, we first report the evaluation results compared to BiTraP using observed and predicted lengths equal to PIE and JAAD.
However, to fairly compare our results on the TITAN dataset with prior work of Malla \textit{et al.}~\cite{malla2020titan}, Social-LSTM~\cite{slstm}, and Social-GAN~\cite{Gupta2018SocialGS}, we follow the same experimental setup used in \cite{malla2020titan}. To that end, we retrain both the BiTraP baseline model, and our proposed model (ABC+) to predict 20 frames, after observing 10 frames, and we report our results using ADE and FDE in pixels, not in squared pixels. 
\subsection{Trajectory Prediction Results} 
The evaluation results are shown in Tab.~\ref{table:pie_jaad} for PIE and JAAD. For TITAN, in Tab.~\ref{table:titan1} we show the evaluation results when observing 10 frames and predicting 20 frames, similar to~\cite{malla2020titan}, and in Tab.~\ref{table:titan2} we compare to BiTraP when observing 5 frames and predicting 10 frames. 
As the tables show, our method (ABC+) achieves superior performance compared to the baseline BiTraP, which does not use our action-based contrastive loss. This result highlights the effectiveness of adding the proposed contrastive objective and sampling strategy. Our proposed method also outperforms other baseline methods, with significant margins.

These evaluation results confirm the gains obtained by using our proposed action-based contrastive loss and sampling strategy. Utilizing action information with our contrastive approach exhibits improved performance across all evaluated benchmarks. In the TITAN dataset, particularly, the performance benefits appear larger. We believe this is due to TITAN's more comprehensive action class structure, compared to PIE or JAAD. In other words, a more diverse set of pedestrian action classes improves the learned embedding space by our action-based contrastive loss. Nevertheless, our method improves the obtained results even with a simpler binary action class structure in PIE and JAAD. 

Another significant result is that our method (ABC+) outperforms the baseline Titan-AP~\cite{malla2020titan} on TITAN, which incorporates the same action class information with observed trajectory information, and produces a combined embedding to predict the future trajectory. This indicates that our approach of supporting the trajectory prediction model with behavioral context information by using action-based contrastive loss is more effective than encoding the action classes in the embedding space representation.  


\begin{table}
\begin{center}
\caption{Ablation results on \textbf{PIE}, \textbf{JAAD}, and \textbf{TITAN} datasets. ABC+ is our proposed action-based contrastive framework with sampling from a learned CVAE. ABC uses our action-based contrastive loss but without sampling from a learned CVAE. SimCLR uses a normal batch contrastive loss instead of our proposed action-based contrastive loss. Lower is better.}
\label{table:ablation}
\begin{tabular}{ x{0.7cm} p{1.8cm} x{0.8cm}x{0.8cm}x{0.8cm} x{1.2cm} x{1.2cm} }
\toprule
& \multirow{2}{*}{Method} &  \multicolumn{3}{c}{ADE} & \multicolumn{1}{c}{C-ADE}& \multicolumn{1}{c}{C-FDE} \\
 & & 0.5&1.0&1.5&  1.5&  \\
\midrule
\multirow{3}{*}{\rotatebox[origin=c]{90}{\textbf{PIE}}} 
 & SimCLR & 26 & 67 & 163& 125 & 399\\
 & ABC &  16 & 40 &  93 & 69 & 213\\
\cmidrule{2-7}
 & ABC+ & \textbf{16} & \textbf{38} & \textbf{87} & \textbf{65} & \textbf{191}\\
\midrule
\midrule
\multirow{3}{*}{\rotatebox[origin=c]{90}{\textbf{JAAD}}} 
 & SimCLR & 50 & 124& 273 & 211 & 608 \\
 & ABC  & 41 & 93 & 201 & 150 & 425\\
\cmidrule{2-7}
 & ABC+ & \textbf{40} & \textbf{89} & \textbf{189} & \textbf{145} & \textbf{409} \\
\midrule
\midrule
\multirow{3}{*}{\rotatebox[origin=c]{90}{\textbf{TITAN}}} 
 & SimCLR & 255 & 506 & 999 & 773 &1805\\
 & ABC  & 188 & 345 & 634 & 488 & 951\\
\cmidrule{2-7}
 & ABC+ & \textbf{165} & \textbf{302}  & \textbf{575} & \textbf{434} & \textbf{843} \\
\bottomrule
\end{tabular}
\end{center}
\end{table}

\begin{figure}
\centering
  \includegraphics[width=\linewidth]{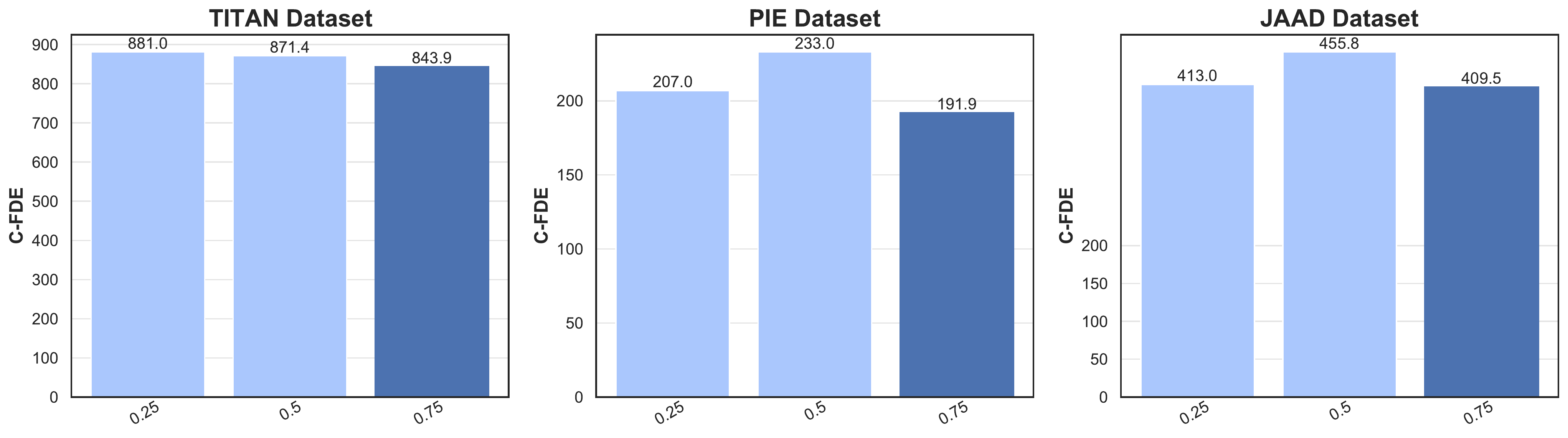}
  
  \includegraphics[width=\linewidth]{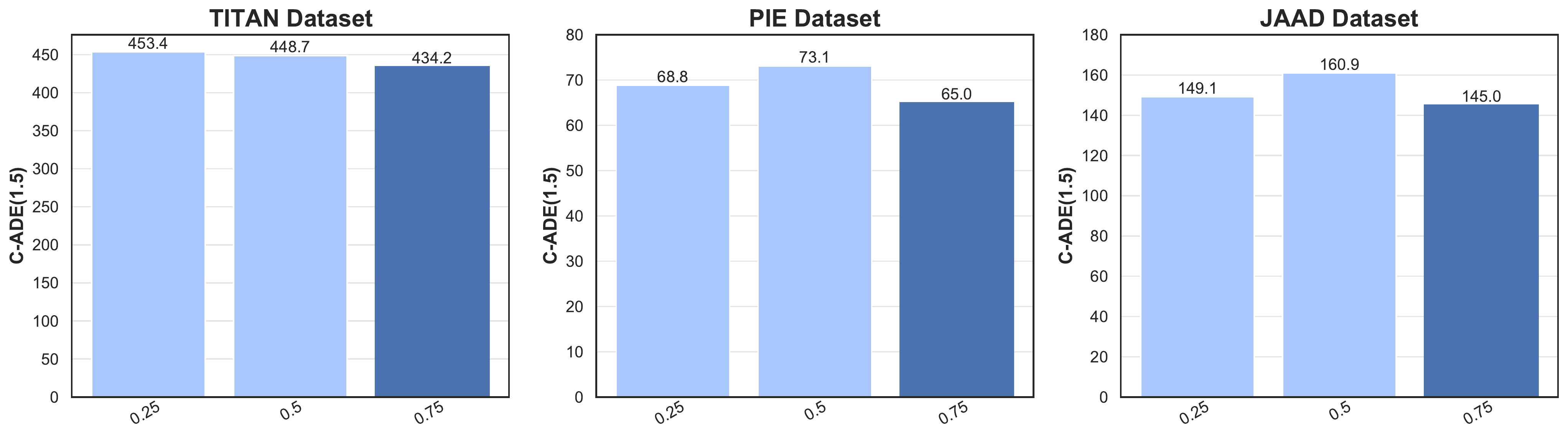}
  \caption{C-FDE results (top) and C-ADE(1.5) (bottom) of trajectory prediction model by applying different  $\beta$ values {0.25,0.5,0.75} in Eq.~\ref{eq_final}. The results are reported for TITAN, PIE, and JAAD datasets. Lower values are better.}
  \label{fig3}
\end{figure}

\subsection{Ablation study}
In this section, we present the ablation studies to provide further insights into our proposed action-based contrastive loss. Similar to the evaluation results shown above, we refer to our proposed action-based contrastive framework by ABC+, where we use the action-based contrastive loss as a regularizer to the trajectory prediction loss, and we also increase negative and positive samples during training by sampling synthetic trajectories from CVAE.

\textbf{Does action information improve the contrastive loss?} The first ablation study examines how the action-based contrastive loss Eq.~\ref{eq_actcon} compares to the batch contrastive loss, namely SimCLR \cite{simclr}, shown in Tab.~\ref{table:ablation}. For this baseline, we replace the action-based contrastive loss with SimCLR contrastive loss, and we measure the trajectory prediction performance. The results demonstrate the impact of utilizing contextual information in form of action on the future trajectory prediction model. 

\textbf{Is the proposed action-based sampling strategy using a CVAE effective?} The second ablation study analyzes the impact of sampling synthetic trajectories from CVAE on the trajectory prediction model performance. The baseline ABC in Tab.~\ref{table:ablation} indicates the trajectory prediction model trained with action-based contrastive loss without using the extra synthetic samples from the learned CVAE. 
Comparing the quantitative results of ABC+ to the results of ABC highlights the effectiveness of our novel sampling strategy on all datasets.

\textbf{How does the weight of the contrastive loss affect the results?} Finally, we also study the influence of the hyper-parameter $\beta$ in Eq.~\ref{eq_final}, which controls the impact of the action-based contrastive loss into the final objective function. As shown in Fig.~\ref{fig3}, we obtain the best results on all benchmark datasets by setting $\beta$ to 0.75.

\section{Discussion and Conclusions}
We presented a contrastive framework for learning behavior-aware pedestrian trajectory representations. Our proposed framework consists of an action-based contrastive loss, and a novel trajectory sampling technique from a learned distribution of a C-VAE model. 
The proposed framework significantly improves the performance of trajectory prediction models on three different first-person view benchmarks. Our evaluation results provide evidence that including pedestrian behavior information, in the form of action or activity class in this case, is beneficial for trajectory prediction. 
Moreover, our results also confirm that our action-based contrastive loss, in conjunction with our sampling strategy, is superior to alternative approaches that also utilize action class information. 

This work comes with a number of strengths. First, we ensure our proposed contrastive loss can be easily integrated with commonly used trajectory prediction models. Second, our proposed sampling strategy utilizes readily learned distributions by generative models, such as CVAEs, and it avoids designing data-specific heuristics. This allows for wider range of applications, such as on animal trajectory data. Finally, contrastive learning in general, and our proposed action-based framework, in particular, allow for enhancing the quantities of underrepresented action classes in the data. This line of work may help address the shortage of necessary edge-cases in training datasets, which may be encountered in real-world scenarios.

This work also comes with a limitation. While effective, our proposed action-based contrastive framework requires pedestrian action labels during the training phase only. However, this requirement is mitigated by our new trajectory sampling technique from CVAE, which does not require action labels for generated samples. Making the training scheme semi-supervised in our model. 
In addition, many modern trajectory datasets are increasingly providing action information. Action prediction tasks from video data achieve high performances \cite{app11188324}, and hence can be performed efficiently and reliably as a pre-processing step for our trajectory prediction framework. We deem evaluating such idea as future work.
\section*{Acknowledgment}
This work was funded by the Deutsche Forschungsgemeinschaft (DFG, German Research Foundation) under Germany’s Excellence Strategy – EXC 2002/1 “Science of Intelligence” – project number 390523135.
\clearpage
\bibliographystyle{splncs04}
\bibliography{egbib}
\end{document}